\documentclass[11pt]{article}
\usepackage{coling2020}
\usepackage{times}
\PassOptionsToPackage{hyphens}{url}
\usepackage{url}
\usepackage{latexsym}
\usepackage{tikz}
\usepackage{pgfplots}
\usepackage{graphicx}
\usepackage{subcaption}
\usepackage{booktabs}
\usepackage{microtype}

\title{The Language of Food during the Pandemic:\\
Hints about the Dietary Effects of Covid-19}

\author{Hoang Van, Ahmad Musa, Mihai Surdeanu, Stephen Kobourov\\
  Computer Science Department, University of Arizona\\
  {\tt \{vnhh,ahmadmusa,msurdeanu,kstephen\}@email.arizona.edu}}

\date{}

\begin{document}
\maketitle
\begin{abstract}
  %In this work we analyze the 
We study the language of food on Twitter during the pandemic lockdown in the United States, focusing on the two month period of March 15 to May 15, 2020.
%March 15--May 15, 2020. 
Specifically, we 
%perform our analysis by
analyze over 770,000 tweets published during the lockdown and the equivalent period in the five previous years
%. Our analysis 
and highlight several worrying trends. First, we observe that during the lockdown there was a notable shift from mentions of healthy foods to unhealthy foods. Second, we show an increased pointwise mutual information of depression hashtags with food-related tweets posted during the lockdown and an increased association between depression hashtags and unhealthy foods, tobacco, and alcohol during the lockdown. 
\end{abstract}

\section{Introduction}
The severe acute respiratory syndrome coronavirus (SARS-CoV)  emerged in late 2002, and caused an outbreak of severe acute respiratory syndrome (SARS) \cite{petersen2020comparing}. The SARS-CoV-2 virus, which causes the Coronavirus Disease 2019 (Covid-19), is closely related to SARS-CoV. On March 11, 2020, World Health Organization (WHO) declared Covid-19 as a pandemic \cite{who2020characterizes}. Covid-19 has claimed 682,855 lives worldwide as of August 1, 2020\footnote{\url{https://en.wikipedia.org/wiki/Template:COVID-19_pandemic_data}}. 

While the death toll of the disease is horrific, there are other indirect public health effects of Covid-19 that are longer term and harder to measure. For example, the disease has already caused many countries to enter a phase of recession \cite{sulkowski2020covid}, which is likely to also affect individual well being. Community anxiety can rise abruptly during a pandemic like Covid-19 \cite{lima2020emotional}. This paper focuses on the early detection of some of these indirect effects using Twitter.

% ms: obvious
%Nowadays, people use social media platforms to share their thoughts and opinions with their connected ones. Their daily posts leave a shadow of their day to day activities.  Twitter\footnote{\url{https://twitter.com/}} is an online social media network used by millions of people around the world \cite{milstein2009twitter}. 
Like other social media platforms, Twitter\footnote{\url{https://twitter.com/}} can potentially serve as a valuable information resource for various public health applications. Among others, 
the paper \cite{bell2018detecting} proposed a method to detect individuals who are at risk of getting type-2 diabetes 
%mellitus (T2DM) 
by analyzing their tweets. The predictive power of the language of food on Twitter and analyzed changes in this language over time are shown in \cite{van2019does}.
%discovered food trends on Twitter, and analyzed how the language of food changes over time. 
%Previously, it has been shown that social media posts contain useful signals that can predict the rise of depression in individuals 
\cite{de2013predicting} showed that social media posts  contain useful signals that can predict the rise of depression in individuals. Symptoms associated with mental illness are observable on Twitter, Facebook, and web forums \cite{guntuku2017detecting}.

% ms: move to approach
%Pointwise mutual information has been found useful to find trends on twitter \cite{van-etal-2019-language}. Here, we use word count and pointwise mutual information to produce the food and depression trend on twitter during Covid-19. Moreover, the dataset has been splited based on the tweet location to show the consistency of the trends on city level, region level and state level. 

Our work compares the language of food on Twitter during the lockdown period in the U.S. (March 15 to May 15, 2020) against the equivalent months in the five previous years. Our analysis highlights worrying trends in dietary patterns and mental health during this interval, which are, thus, likely to be indirect effects of the disease and the associated lockdown. In particular,
the contributions of this work are two-fold:

\noindent {1.} Using a dataset of 772,142 tweets and a lexicon of 800 good phrases, we show that during the lockdown period there was a significant change in the language of food on the U.S. Twitter. 
In particular, we observe a shift from mentions of healthy foods to unhealthy foods on the U.S. Twitter, compared to previous years. This trend is worrying because previous works showed that the consumption of unhealthy food leads to obesity \cite{karnik2012childhood,chopra2004tobacco}. Further, there is strong evidence that obesity is a significant comorbidity factor for complications from the corona virus \cite{stefan2020obesity}. 
Our result contradicts the general belief that the lockdown led to better diets due to an increase in home cooking,\footnote{See, for example, the Google trend for the word ``sourdough'': \url{https://trends.google.com/trends/explore?q=sourdough&geo=US}.} and evidence from other countries such as Italy \cite{Renzo2020}.
%\fix{I'm not able to understand properly. I think it's related to the google trend of fast food and soourdough. I think Mihai came up with that excellent example.}
%\todo{sk: let's bring back the ``despite the reduction in fast food consumption and increase in home cooking"}
% We count exact matches of healthy and unhealthy food words from the tweets to catch the shift. Moreover, By limiting the data from March 15 to May 15, we maintain the relevance of the experiment to Covid-19. Even though the timespan of our collective tweets is only 6 years, the signals captured by our experiments are significantly strong to indicate the shift. 

\noindent {2.} We analyze the incidence of depression during the lockdown interval using a set depression hashtags such as {\tt \#depression} and {\tt \#suicidalthoughts}. Our analysis %discovered 
shows a considerable increase in pointwise mutual information (PMI) of these depression hashtags and tweets coming from the lockdown interval vs. equivalent months in the previous years. Further, we observe an increased association between depression hashtags and unhealthy foods during lockdown compared to previous years. This analysis suggests that not only was there an increase in depression during the lockdown, but also that this increase led to a larger negative impact on dietary patterns than before the Covid-19 pandemic. 
%Our goal is to use a set of depression hashtags to compute pointwise mutual information with respect to each year to analyze the depression trend on twitter. Also, hashtags are indicators of breaking events \cite{10.1145/2396761.2398519}. Therefore, the filtered dataset with these hashtags is useful to find depression-related food trends during Covid-19. We also report top depression-related food choices over time.

\section{Data} \label{sec:data}

The dataset we collected focuses on food-related tweets.
We used Twitter's public streaming API\footnote{\url{https://developer.twitter.com/en/docs/tweets/filter-realtime/guides/connecting.html}} to collect tweets and their metadata such as geolocation and temporal information. We filtered the tweets using a list of seven meal related hashtags (Table \ref{tab:tweetdataset}). Tweets have been stored into a Lucene-backed Solr instance,\footnote{\url{https://lucene.apache.org/}. Solr is the open-source NoSQL search platform from the Apache Lucene project.} which was used to localize the tweets within the U.S. In total, we have collected 28 million tweets from the period between October 2, 2013 and May 25, 2020. Similar to \cite{van2019does}, we localized 5 million of these tweets to a U.S. location using either the geo tag associated with the tweet, or the user's self-reported location.

From the above dataset, we extracted tweets that were localized in the U.S., contained a meal-related hashtag, and were posted between March and May from year 2015 to 2020. This smaller dataset is summarized in Table~\ref{tab:tweetMonthly}. This is the dataset we used for the analyses reported in this paper.
Table~\ref{tab:tweetMonthly} indicates that the number of meal-related tweets decreased from 185K in March--May 2015 to only 63K in the same period in 2020. This decrease may be explained by multiple overlapping factors: (a) the overall number of tweets is decreasing;\footnote{\url{https://www.businessinsider.com/tweets-on-twitter-is-in-serious-decline-2016-2}} (b) the number of tweets that are geo tagged is decreasing due to an increasing user focus on privacy;\footnote{\url{https://www.forbes.com/sites/kalevleetaru/2019/03/04/visualizing-seven-years-of-twitters-evolution-2012-2018/}} and (c) lastly, we suspect that the meal-focused social media is migrating to platforms that are more multi-modal friendly such as Instagram. Nevertheless, the overall counts are sufficient to draw reliable conclusions.

\begin{table*}[t]
\centering
\scalebox{1.0}{%
 \begin{tabular}{c c c} 
 \toprule
Term & \# of tweets & \# of tweets localized in U.S. \\
 \midrule
 \#dinner & 6,007,037 & 1,520,573 \\
 \#breakfast & 5,736,525 & 1,335,432 \\
 \#lunch & 5,514,638  & 1,213,171 \\
 \#brunch & 2,160,015 & 768,577 \\
 \#snack & 898,178 & 246,842 \\
 \#meal & 553,214 & 115,101 \\
 \#supper & 138,658 & 24,860 \\
 \midrule
 Total & 28,854,023 & 5,198,387 \\
 \bottomrule
\end{tabular}}
\caption{Seven meal related hashtags and corresponding number of tweets  in the complete tweet dataset, from October 2013 to May 2020. The right-most column indicates the number of tweets we could localize to a U.S. state or Washington D.C.
\label{tab:tweetdataset}}
\end{table*}

\begin{table*}[t]
\centering
\scalebox{1.0}{%
 \begin{tabular}{c c} 
 \toprule
Year (March-May) & \# of tweets localized in U.S. \\
 \midrule
 2020 & 63,898 \\
 2019 & 81,036  \\
 2018 & 131,418 \\
 2017 & 135,734 \\
 2016 & 174,792  \\
 2015 & 185,264 \\
 \bottomrule
\end{tabular}}
\caption{Number of tweets containing meal-related hashtags, which were localized in the U.S. and were posted in the March--May interval during the six years analyzed.\label{tab:tweetMonthly}}
\end{table*}

\begin{figure*}[t!]
\centering
% ms: 100% of linewidth is too large
\includegraphics[width=1.0 \linewidth]{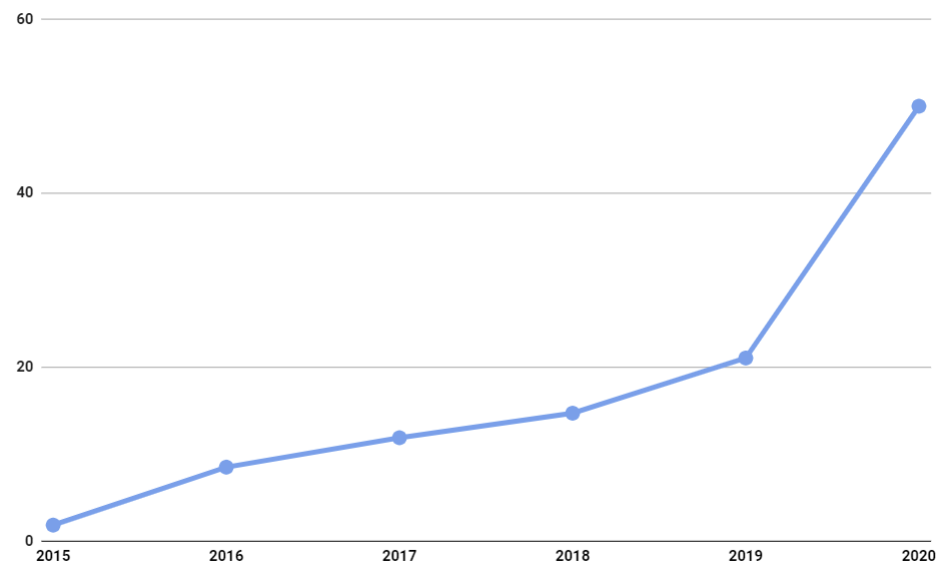}
\caption{\label{fig:delivery_pmi} PMI values for delivery service hashtags (e.g., \#ubereat, \#grubhub, \#doordash, \#postmates)  over the six years investigated.}
% Here, all delivery service hashtags are treated as a single term. 
%Increases in PMI values indicate an upward trend for the mentions of delivery service hashtags. This trend highlights that delivery service hashtags are on an upward trend, especially during the Covid-19 lockdown.}
\vspace{-2mm}
\end{figure*}

\section{Approach}

The two main aims of this work are: (a) to explore the change in dietary patterns mentioned in these tweets during the Covid-19 lockdown period, and (b) to identify the change in mental health trends in the same period, and their associations with diet. We discuss the settings of these experiments next. 
%there was a shift from mentions of healthy to unhealthy foods in tweets under effects of the Covid-19 pandemic lockdown and the its side effect to mental health such as depression through the related choice of food.  

\subsection{Food trends in tweets}
We investigate dietary trends (as mentioned in tweets) through three distinct analyses that range from nation-level, to region-level, to  city-level. For city-level data, we analyze the changes in food trends in New York City, Chicago, and Los Angeles, which are the top three cities most affected by the  pandemic~\cite{dong2020interactive}. We use the tweets from the above dataset as the only input for our analyses. To show the effect of the Covid-19 pandemic lockdown, we compare tweets in the interval March 15 to May 15 between the years 2015 to 2020.

\begin{table*}
\centering
\scalebox{1.1}{\begin{tabular}{c}
\toprule
\#depression   \#depress \#depressed    \#stressed     \#anxiety \#anxious \#sicknotweak \#sadness \\
\#sucidalthoughts \#selfharming  \#feelingdown \#bipolar \#mentalhealthawareness \#ocd \\
\#mentalillness	\#anxietyrelief	\#depressionawareness \#mentalhealth \#itsokaynottobeokay \\ \#stopthestigma \#stress \#mentalhealthsupport \#mentalhealthrecovery  \#killyourself \\
\#yourstoryisnotover \#recoveryisworthit \#recoveryispossible
 \#socialanxiety  \\
\#schizophrenia	\#killyourself	\#kindnessmatters \#suicideprevention \#ptsd \#suicidal\\
\bottomrule
\end{tabular}}
\caption{List of hashtags related to depression used in our experiments.}\label{tab:depression_hashtags}
\end{table*}

{\flushleft{\bf Nation-level food trends:}} For each tweet localized in the U.S., we count the number of matches of healthy and unhealthy food words from the dictionary included in \cite{van2019does}. 
This dictionary contains a food lexicon, with foods organized in three categories: healthy, unhealthy, and neutral. Table \ref{tab:food_category_examples} shows several examples from this dictionary.
We normalize the food word counts collected from tweets by the overall number of food words observed per year.
%To investigate if people tweet more about unhealthy or healthy foods, we calculate the percentage of the total occurrences of food words in each food category over the total occurrences of all the food words across all the localized tweets described in Section \ref{sec:data}.  

\begin{table*}[h!]
\centering
\scalebox{1.0}{%
\begin{tabular}{p{1.5cm}|p{9cm}}
\toprule
Healthy & peach, blueberry, succotash, whey, acorn, cabbage, mushroom, beans, beetroot, banana \\
\midrule
Unhealthy & quesadilla, ham, wine, beer, cake, soda, liquor, milkshake, hamburger, cheese  \\
\midrule
Neutral & crunch, ketchup, drink, mussels, fish, meat, soup, salt, sandwich, protein  \\
\bottomrule
\end{tabular}}
\caption{Examples for each food category in the food vocabulary in \cite{van2019does}.   \label{tab:food_category_examples}}
\end{table*}

{\flushleft{\bf Region-level food trends:}} We use the region division by the U.S. Census Bureau,
%\footnote{https://www2.census.gov/geo/pdfs/maps-data/maps/reference/us_regdiv.pdf} 
which identifies the following 4  regions: Northeast, Midwest, South, and West. 
We analyze the data in a way similar to that for the nation-level, but focusing on each of the regions individually. 
% previous discussion,
We count mentions of healthy/unhealthy food words using the dictionary of \cite{van2019does}, and normalize these counts by the total number of food words in each region. 
%For each tweet localized in the four regions of the U.S., we count exact matches of food words in healthy and unhealthy food dictionaries in \cite{van-etal-2019-language}. To investigate if people tweet more about unhealthy or healthy food in each region, we calculate the percentage of the total occurrences of food words in each food category over the total occurrences of all the food words across all the tweets localized in that region.

{\flushleft{\bf City-level food trends:}} We analyze the food trends in New York City, Chicago, and Los Angeles, which are the top three cities most affected by the Covid-19 lockdown \cite{dong2020interactive}. Again, we use the healthy/unhealthy food word dictionary, and normalize counts by the total count of food words in each city. 
%For each tweet localized in the three U.S. cities, we count exact matches of food words in healthy and unhealthy food dictionaries in \cite{van-etal-2019-language}. To investigate if people tweet more about unhealthy or healthy food in each city, we calculate the percentage of the total occurrences of food words in each food category over the total occurrences of all the food words across all the tweets localized in that city.

\subsection{Mental health effects of the lockdown}
Previous work has shown that the Covid-19 pandemic lockdown can worsen mental health and cause mental problems such as anxiety and depression \cite{kumar2020covid,torales2020outbreak,gualano2020effects}. To complement this work, we analyze mentions of hashtags related to mental health problems in tweets during the Covid-19 lockdown, compared against the same period of the previous 5 years. We provide three analyses for this goal: trends for depression-related hashtags, associations between these depression-related hashtags and food words, and the top depression-related food choices over time.

{\flushleft{\bf Trends of depression-related hashtags:}} To analyze these trends over time, we measure the pointwise mutual information (PMI) between depression hashtags (see Table \ref{tab:depression_hashtags}) and tweets in the time period corresponding to the lockdown vs. the same period in previous years. 
In particular, we divide our corpus into six parts, each containing the complete set of tweets between March 15 and May 15 of the corresponding year (2015--2020), and then calculate the PMI for pairs (depression hashtags, year $y$) using the formula:
\begin{equation} \label{eq:pmi}
    PMI(t, y) = \frac{C(t,y)}{C(t)*C(y)}
\end{equation}
where $C(t,y)$ is the number of tweets with depression hashtags in the period corresponding to year $y$, $C(t)$ is the total number of tweets with depression hashtags across all years, and  $C(y)$ is the total number of tweets in the period corresponding to year $y$. 
% ms: fixed
%\todo{Are you counting hashtags or tweets here? You should count one or the other but not both. You seem to count hashtags for C(t, y) and C(t), and tweets for C(y), which is wrong! Hi Mihai, I meant co-occurrence of tweet and depressions, or in other words C(t,y) is the tweets in year y contains depression hashtags, C(t) is a count of tweets with depression hashtags across all years and C(y) is total of tweets in year y}
%Here, all depression hashtags are treated as a single term. % ms: unclear
Intuitively, the higher the PMI value of a term in a given year, $PMI(t, y)$ the more that term is associated with tweets from that year in particular.

\begin{table*}[t]
\centering
\scalebox{1.0}{%
 \begin{tabular}{c l c c c c c} 
 \toprule
 \# & \textbf{Year} & \textbf{U.S} & \textbf{Northeast} & \textbf{Midwest} & \textbf{South} & \textbf{West} \\
 \midrule
 1 & 2020 & 70.23 & 70.67 & 71.38 & 67.50 & 70.17 \\
 \midrule
 2 & 2019 & 60.66 & 59.65 & 65.35 & 60.52 & 58.99 \\
 \midrule
 3 & 2018 & 62.04 & 61.71 & 66.99 & 62.53 & 61.13 \\
 \midrule
 4 & 2017 & 65.31 & 64.32 & 68.55 & 65.34 & 64.91 \\
 \midrule
 5 & 2016 & 62.41 & 62.00 & 63.86 & 62.77 & 60.83 \\
 \midrule
 6 & 2015 & 61.48 & 60.17 & 62.93 & 61.01 & 59.36 \\ 
 \bottomrule
\end{tabular}}
\caption{ Percentage of unhealthy foods mentioned in tweets during the interval March 15 -- May 15 in each year analyzed. The U.S. column lists the percentage of mentions of unhealthy foods nation wide. The Northwest, Midwest, South, and West columns list the percentages of mentions of unhealthy foods for the four U.S. regions. \label{tab:region_percentage}}
\end{table*} 

{\flushleft{\bf Depression-related foods trends:}} For this analysis, we investigate only the U.S. localized tweets that contain depression hashtags. We compute PMI values between food words and tweets that contain depression hashtags (again, considering just the March 15 -- May 15 interval) to analyze the association between healthy/unhealthy foods and depression over time. The PMI for all pairs (food $t$, year $y$) is calculated using formula \ref{eq:pmi}, where $C(t,y)$ is the number of depression-related tweets from year $y$ containing food $t$, $C(t)$ is the total number of depression-related tweets with food $t$ across all years, and $C(y)$ is the total number of depression-related tweets in year $y$.
% ms: fixed
%\todo{again, this is wrong: you are combining hashtag counts and tweet counts!Hi Mihai, same for above, I meant co-occurrence, which is in other word the tweets.}. 
%Here, all healthy foods are treated as a single term. The same thing goes for unhealthy foods. 

{\flushleft{\bf %Top
Depression-related food choices:}} We investigate only the U.S. localized tweets that contain depression hashtags, and rely on the PMI values computed in the previous step.
% ms: redundant!
%We use PMI between food words and years to analyze top food words associated with depression over time. We divide our corpus into six parts, each contains complete year's set of depression-related tweets in March-May period (2015-2020) and then calculate PMI for pairs (food t, year y) using formula \ref{eq:pmi}, where $C(t,y)$ is the number of occurrences of food t in year y, C(t) is the total number of occurrence of food t, C(y) is the total number of depression-related tweets in the year y. To further understand the change in food choices during the Covid-19 pandemic lockdown, 
Here, we identify the list of foods with largest changes in PMI from previous years. We divide our corpus into two parts, the first part contains tweets in the March--May period in 2020 and the second part contains tweets in the same period in {\em all} previous years in our dataset (2015--2019). We then calculate the differences in PMI for each food words between two partitions, and report foods with the largest differences.

\section{Results and Discussion}
Next we discuss the results of the analyses introduced in the previous section.
% ms: redundant
%We present the results for all analyses of the shifts in eating habits over time (i.e., food trends) at nation, region, and city levels. We also investigate the depression trends in tweets and the significant changes in related food choices over the years.

\subsection{An unhealthy shift in dietary trends}

{\flushleft{\bf Nation-level food trend:}} Table \ref{tab:region_percentage} lists the percentage of unhealthy foods mentioned in tweets (normalized by the total number of food words in the corresponding interval). The table shows that the largest increase at the nation level in unhealthy food mentions from the previous year occurred in 2020, during the Covid-19 pandemic lockdown (a relative increase of 15.77\%). 
The lockdown period in 2020 had the largest total percentage of unhealthy food mentions, at 70.2\%. 

Interestingly, the second highest percentage of mentions of unhealthy foods in the March--May period occurred in 2017 (a total of 65.3\% with a relative increase of 4.65\% from 2016) when the U.S. experienced a seismic shift in national politics. The next two years after 2017 saw a downward trend in mentions of unhealthy foods, but this trend was abruptly reversed in 2020.

This analysis suggests that the lockdown period in 2020 had a notably negative effect on American diets. This is a concerning observation, as dietary patterns are in important indicators of public health~\cite{gorski2015public}. 
\begin{figure*}[t!]
\centering
% ms: 100% of linewidth is too large
\includegraphics[width=1.0 \linewidth]{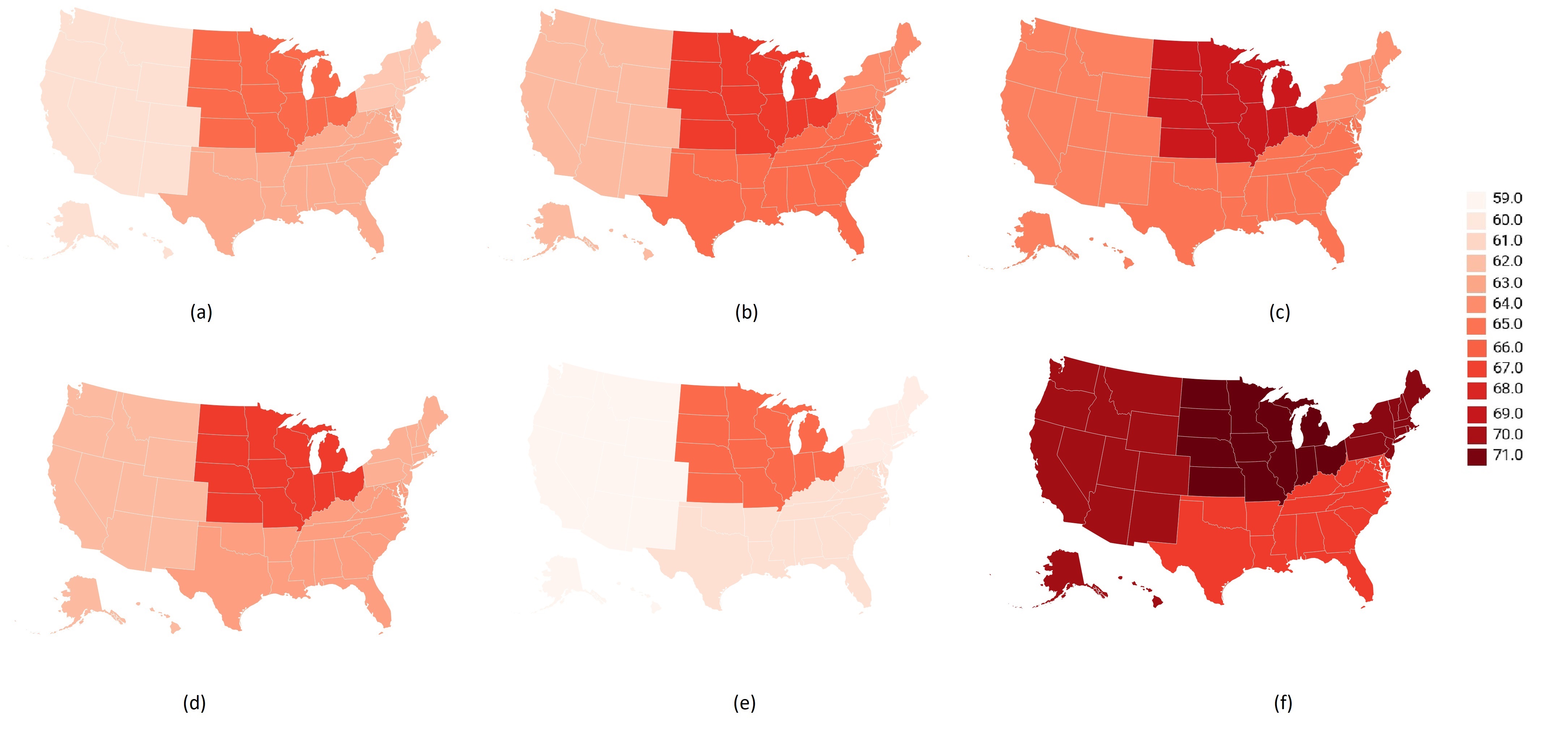}
\caption{\label{fig:PercentageUnhealthy} A series of U.S. regional maps, where the (a)--(f) maps plot the years 2015 to 2020, respectively. Regions with higher percentage of unhealthy food tweets (posted in the March--May period each year) have darker red color. The year 2020 shows a clear peak over the previous five years.}
\vspace{-2mm}
\end{figure*}

\begin{figure*}[t!]
\centering
% ms: 100% of linewidth is too large
\includegraphics[width=1.0 \linewidth]{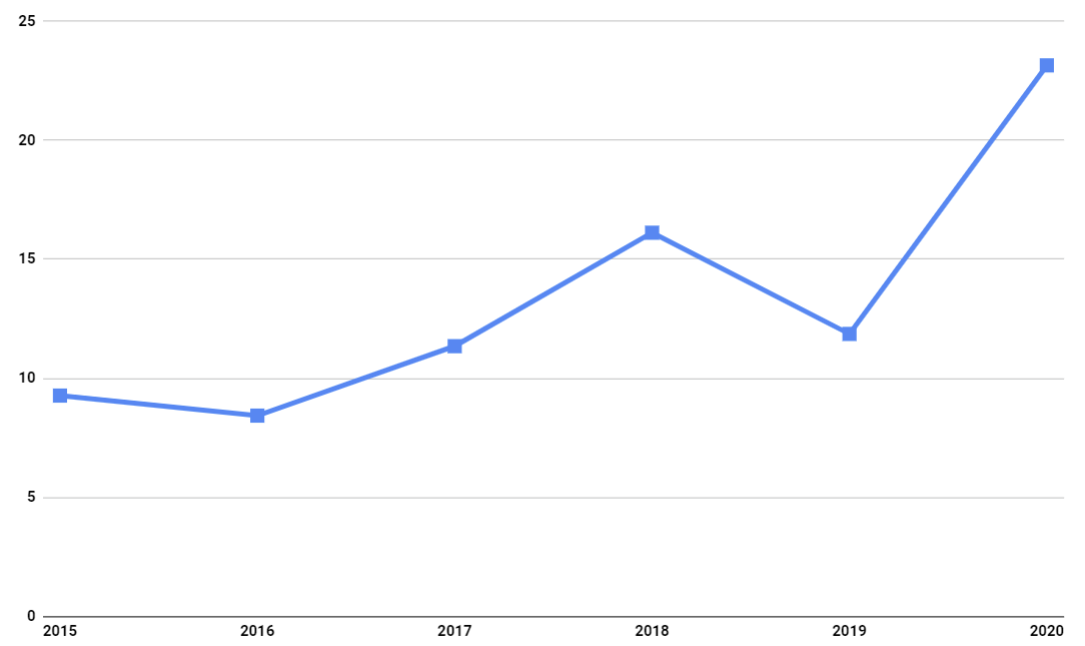}
\caption{\label{fig:depression_pmi} PMI values for depression hashtags over the years.}
%Here, all depression hashtags are treated as a single term.
%Increases in PMI values indicate an upward trend for the depression hashtags. This trend highlights that depression hashtags are on an upward trend, especially during the Covid-19 lockdown.}
\vspace{-2mm}
\end{figure*}

To further highlight this negative trend, we also investigated the number of mentions for delivery service hashtags over time.
This analysis is motivated by the strong association between online food delivery services and unhealthy eating habits \cite{cetateanu2014understanding}. Figure \ref{fig:delivery_pmi} shows an upward trend for delivery service hashtags over the six years investigated. (Intuitively, the higher the PMI value of a term in a given time period the more that term is associated with tweets from that period.) However, the PMI score for delivery service hashtags during the Covid-19 pandemic lockdown increased drastically more than the trend observed in previous years, more than doubling its value compared to the same period in 2019.

\begin{figure*}[t!]
\centering
% ms: 100% of linewidth is too large
\includegraphics[width=1.0 \linewidth]{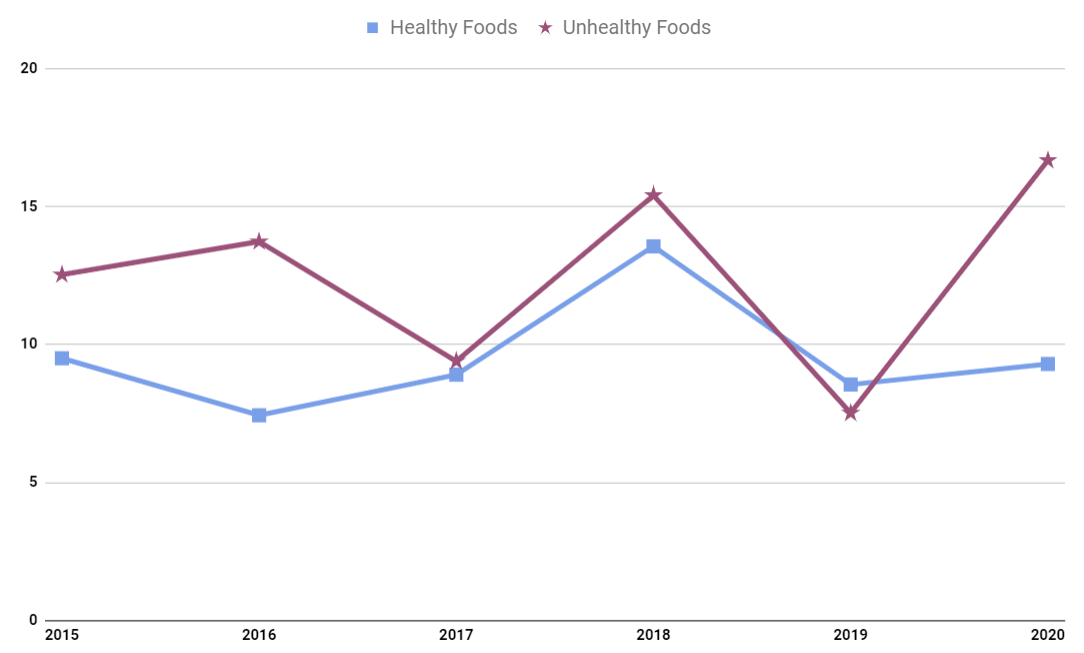}
\caption{\label{fig:depression_healthy_unhealthy_pmi} PMI values for each food category (healthy and unhealthy) over the six years, for depression-related tweets. }
%Here, all foods in one category, e.g. healthy, are treated as a single term. Increases in PMI values indicate an upward trend for the food categories.}
\end{figure*}

%From 2015 to 2017, there is an upward trend in mentions of unhealthy foods, i.e., an increase of 3.83\% in a two year period \todo{unclear. what do you mean by "two year period"? please explain}. In another two year from 2017 to 2019, there is an downward trend for unhealthy foods, i.e. a decrease of 4.65\%. However, during the Covid-19 lockdown, there is an acute increase of 9.57\% in mentions of unhealthy food. This proves that during the lockdown, people tend to consume more unhealthy foods and the lockdown has negatively impacted the eating habits which is an important aspect of the public health \cite{gorski2015public}. Besides analyses of strong increases in mentions of unhealthy foods, to show there is an unhealthy shift in food trends across the U.S., we also investigate the trend for delivery service hashtags over time because there is a strong relation between online food delivery services and unhealthy eating habits \cite{cetateanu2014understanding, harahap2020relationship,love2020go}. Figure \ref{fig:delivery_pmi} shows a clear upward trend for delivery service hashtags (i.e. \#ubereat, \#grubhub, \#postmates, \#doordash). The PMI score for delivery service hashtags is doubled during the Covid-19 pandemic lockdown. Intuitively, the higher the PMI value of a term in a given time period the more that term is associated with tweets from that period in particular.

{\flushleft{\bf Region-level food trends:}} The Northeast and West regions of the U.S. were the most affected by the Covid-19 pandemic \cite{dong2020interactive}. The two regions were the first to experience state-wide lockdowns. Table \ref{tab:region_percentage} reflects these impacts by the Covid-19 lockdown on the regional level. The Northeast and West regions have the largest relative increases in unhealthy foods, with 18.47\% and Hoang 18.95\%, respectively. These increases are well above the increase in unhealthy food mentions at country level. The Midwest and South regions also experience large relative increases in mentions of unhealthy foods (9.23\% and 11.53\%) from previous years, but these increases are below  the national average.
Figure~\ref{fig:PercentageUnhealthy} summarizes these trends visually, highlighting the worrying increases in 2020. These sharp increases are worrying because the US already has very high obesity rates (39.8\% of adults aged 20 and over are obese and that another 31.8\% were overweight \cite{fryar2018prevalence}). Further, there is strong evidence that obesity is a significant comorbidity factor for complications from the corona virus \cite{stefan2020obesity}.
% ms: statement too strong. not sure we can prove the severity of a lockdown
%This finding further confirms that the more severe the lockdown is, the stronger the effects caused by the Covid-19. 

\begin{table*}[t]
\centering
\scalebox{.9}{%
 \begin{tabular}{c l c c c} 
 \toprule
 \# & \textbf{Year} & \textbf{New York} & \textbf{Chicago} & \textbf{Los Angeles} \\
 \midrule
 1 & 2020 & 72.01 & 73.27 & 71.68  \\
 \midrule
 2 & 2019 & 61.61 & 64.39 & 61.02  \\
 \midrule
 3 & 2018 & 63.56 & 66.74 & 62.37  \\
 \midrule
 4 & 2017 & 65.74 & 67.22 & 63.45  \\
 \midrule
 5 & 2016 & 63.51 & 66.69 & 63.86  \\
 \midrule
 6 & 2015 & 61.08 & 64.17 & 63.22  \\ 
 \bottomrule
\end{tabular}}
\caption{ Percentage of unhealthy foods mentioned in tweets in New York City, Chicago, and Los Angeles over the six years analyzed.\label{tab:city_percentage}}
\end{table*}

{\flushleft{\bf City-level food trends:}} To conclude the impacts of the Covid-19 pandemic lockdown on dietary habits, we also investigate the unhealthy food trends in tweets in the three most affected cities by the Covid-19 (based on both number of cases and deaths) in the U.S.: New York City, Chicago, and Los Angeles \cite{dong2020interactive}. Table \ref{tab:city_percentage} shows the prevalence of unhealthy foods in the three cities. It is notable that the number of mentions of unhealthy foods in tweets in these three cities is well above both the national average as well as the average increase in the regions to which they belong.

\subsection{Mental health effects of the Covid-19}

{\flushleft{\bf Depression trend:}} Figure \ref{fig:depression_pmi} plots the depression PMI values for the 2015--2020 period. In general, the higher the PMI value, the stronger the association between depression hashtags and tweets from that period is. 
The figure shows that there is an upward trend for depression PMI values from 2015 to 2020, with a sharp increase during the Covid-19 pandemic lockdown in 2020.
Further, the PMI score for depression hashtags in 2020 is the highest across the six periods investigated.
This analysis suggests that the Covid-19 lockdown had a negative impact on the mental health of people in the U.S. Our observation aligns with other measurements of depression during the lockdown, which indicated that depression increased during the lockdown \cite{gualano2020effects,fullana2020coping,smith2020factors}. 

\begin{table*}[h!]
\centering
\scalebox{.9}{%
\begin{tabular}{p{0.9cm}|p{15cm}}
\toprule
2020 & wine, beer, cigarette, quesadilla, liquor, cocktail, succotash, milkshake, blueberry, citron \\
\midrule
2019 & quesadilla, peach, prune, blueberry, succotash, whey, acorn, ham, burrito, mushroom \\
\midrule
2018 & quesadilla, tomatillo, oyster, yolks, seafood, liquor, lychee, pate, broccoli, rum \\
\midrule
2017 & scallion, quesadilla, nectarine, liquor, beer, pancake, brandy, sirloin, ham, dragonfruit \\
\midrule
2016 & quesadilla, tomatillo, nectarine, maize, pumpkin, ham, soda, coriander, tequila, guacamole \\
\midrule 
2015 &  lychee, dragonfruit, quesadilla, tequila, coriander, endive, lard, daikon, guacamole, seaweed \\
\bottomrule
\end{tabular}}
\caption{Top food words associated with depression hashtags for the March 15--May 15 period over six years, according to their PMI values. The words are listed in descending order of PMI values from left to right.  \label{tab:top_pmi_food_words}}
\end{table*}

{\flushleft{\bf Depression-related foods trends:}} 
We next investigate whether depression is associated with mentions of healthy or unhealthy foods in our data. Figure \ref{fig:depression_healthy_unhealthy_pmi} plots the PMI values of healthy and unhealthy foods over the six time periods analyzed, with respect to tweets that contain at least one depression hashtags. Intuitively, these values indicate the association strength between depression hashtags and healthy/unhealthy food words. Between 2015 and 2019 there are no clear differences between the PMIs of healthy and unhealthy foods. However, during the Covid-19 lockdown, the PMI of unhealthy food words increases sharply, reaching the highest value measured in our dataset. Further, this value is 78\% higher than the PMI of healthy foods in the same period. This suggests that during the pandemic lockdown not only did the incidence of depression increase, but also that it is associated with an increased consumption of unhealthy foods.

{\flushleft{\bf %Top 
Depression-related food choices:}} To further analyze the association between depression and food, we extracted the food words with highest PMI values in depression-related tweets over the six periods investigated. Table \ref{tab:top_pmi_food_words} shows the top 10 food words per year. 
Unsurprisingly, the majority of these top 10 food words are unhealthy. In years when the depression PMI peaks (2018 and 2020), unhealthy foods (i.e., liquor, beer, cigarette, rum, wine) are among the top food choices. This is especially clear during the Covid-19 pandemic, when five out of the top 10 food words are related to alcohol and cigarettes. Several examples of such tweets, which contain mentions of depression and dietary items, are shown in Table \ref{tab:tweet_examples}. 

\begin{table*}[h!]
\centering
\scalebox{.95}{%
\begin{tabular}{p{1.2cm}|p{.3cm}|p{14cm}}
\toprule
2020 & * & During “these \#stressful times”, my \#breakfast is NOW: \#coffee, \#cigarettes, AND \#beer.. I work from home. \\

& * & \#cigarettes \#depression \#breakfast https://t.co/3mJVHXD \\

& * & cigarettes is a new \#dinner \#Covid19 \#mentalhealthmatters \\

& * & \#mentalhealthsupport weed or cigarettes no \#dinner or \#breakfast \\

& * & RT @GPasadena: so pretty! healing \#depression \#Paadena @Madeline\_Garden \#winetasting \#champagnebrunch \#brunch \#sundaybrunch \#breakfast…
 \\
\midrule
Previous Years & * & Time for some \#stressrelief with a glass of crisp \#whitewine! \#LuigiPizzaPasta \#CampbellCA \#Dinner \#mentalhealthmatters https://t.co/JQ2N6Lr \\

& * & RT @tsadok03: Dinner turned to be duck rillette, Croustade de Canard, great chess selection and amazing wines. \#dinner \#food \#wine \#family \#mentalhealthsupport… \\

& * & RT @ThePitmistress: @TomHixsonMeat @snakeriverfarms Umm steak for lunch \#srf \#webber \#foodie \#bbq \#steak \#bunchofswines \#tastesogood \#anxietyrelief… \\

& * & \#wine \#countrybreakfast \#kenwood \#foodtruck \#postup \#breakfast \#eggsadobe \#paleo \#kindnessmatters https://t.co/ASKYKj https://t.co/RqCVFkI \\

& * & Ordering \#wine with \#dinner at your favorite \#restaurant doesn't have to be \#stressful! https://t.co/Bq6j1b04 \\
\bottomrule
\end{tabular}}
\caption{Examples of tweets containing depression hashtags in two periods: the March--May interval in 2020 and the same interval in the
five previous years, highlighting a worrying attitude change in depression-related tweets towards unhealthy foods, tobacco, and alcohol. \label{tab:tweet_examples}}
\end{table*}

\begin{table*}[h!]
\centering
\scalebox{1.1}{%
\begin{tabular} {c}
\toprule
cigarette, wine, vodka, whiskey, tarragon, citron, \\
beer, batter, cocktail, brisket, grapefruit, liquor, \\
fava, jicama, shallots, lamb, lentils, flan, squid, tripe \\
\bottomrule
\end{tabular}}
\caption{Top 20 food words with largest changes in PMI relative to depression-related tweets in 2020 compared to the previous five years. The words are listed in descending order of this value, from left to right. \label{tab:top_change_pmi_food_words}}
%We calculate the differences between PMI scores for food words during the Covid-19 pandemic and all previous years. From left to right, the changes in PMI decrease. Cigarette and liquor-related products appear in the top 10 words with largest positive changes in PMI. Intuitively, the higher the PMI value of a term in a given time period the more that term is associated with tweets from that period in particular.\label{tab:top_change_pmi_food_words}}
\end{table*}

Lastly, Table \ref{tab:top_change_pmi_food_words} shows the top 20 food words that had the largest increase in PMI relative to depression-related tweets in 2020, compared to all previous years.  
%with the positive largest changes in PMI. Intuitively, the higher the PMI value of a term in a given time period the more that term is associated with tweets from that period in particular. 
The top foods in the table are cigarette, wine, vodka, and whiskey, which further confirms a concerning increase in the association between depression hashtags and unhealthy dietary patterns in social media.
%Based on the two analyses, we show that people tend to use unhealthy products such as alcohols and cigarettes as a means to resolve potential mental problems caused by the Covid-19 lockdown.

\section{Conclusion}

In this work we compared the language of food during the pandemic lockdown period in the United States against the same period in five previous years. Our analysis indicates that during the lockdown period there was a considerable shift towards mentions of unhealthy foods compared to previous years. Further, we showed that there was a considerable increase in PMI between depression hashtags and tweets posted during the lockdown. Lastly, we highlighted an increased association between depression hashtags and unhealthy foods, alcohol and cigarettes. 

While all these results are worrying, we did not prove yet that this social media analysis correlates with real life information. We leave this parallel analysis to future work. However, given the previous work that showed that real life public health information can be forecast from social media \cite{van2019does,bell2018detecting,guntuku2017detecting,de2013predicting}, we envision that our analysis can be used as a near real-time monitoring tool for the rapid identification of important public health factors such as diet and mental health. 

% We showed that the Covid-19 pandemic lockdown has negatively affected eating habits and worsened mental health in the U.S.. During the Covid-19 lockdown, there is a strong shift form healthy to unhealthy eating habits. The more severe the lockdown is, the stronger the shift. Further, we showed that there is an alarming upward trend for unhealthy food with respect to mental health problems such as anxiety and depression as side effects caused by the Covid-19 pandemic lockdown. With analyses of an unhealthy shift in food choices and trends of depression-related hashtags, we proved that the language of food can provide real-time and efficient tools in monitoring public health where resource is scarce and quick turnaround time is required, i.e. during the Covid-19 pandemic. 

\bibliographystyle{coling}
\bibliography{coling2020}

\end{document}